\PassOptionsToPackage{section}{placeins}
\documentclass{fairmeta}
\usepackage[utf8]{inputenc}
\usepackage{amsmath,amssymb}
\usepackage{tabularx,array}
\hypersetup{pdftitle={Not Every Rollout Needs Distillation: Correctness-Filtered On-Policy Self-Distillation}}
\title{Not Every Rollout Needs Distillation: Correctness-Filtered On-Policy Self-Distillation}
\author[1]{Harrison (Zihao) Zhao}
\author[1]{Lei Huang}
\author[1]{Zhiyuan Zhao}
\author[1]{Fei Liu}
\author[1]{Wen Shi}
\author[1]{Zijian Jin}
\affiliation[1]{Meta}
\hypersetup{pdfauthor={Harrison (Zihao) Zhao, Zijian Jin}}

\abstract{
On-policy self-distillation (OPSD) supervises student-generated trajectories with a reference-conditioned version of the same base model. We ask whether every trajectory supplies distinct supervision that warrants its distillation cost. We measure gradients on Qwen3-1.7B at initialization and find that correct and incorrect responses have similar individual gradient magnitudes, but incorrect responses have approximately three times the coherent gradient share. Their estimated population-mean directions also largely overlap. A signal--noise model fitted at a small group size tracks alignment at larger group sizes without refitting. Motivated by these measurements, we propose correctness-filtered OPSD (C-OPSD), which generates one response per problem and distills only a subset of incorrect responses. Our method removes unselected rows before both distillation forwards, avoiding their teacher/student forward and student backward work. Across Qwen3-1.7B, 4B, and 8B on AIME24, AIME25, and HMMT25, our method matches the performance of full OPSD. It executes approximately 68\% fewer distillation rows and, with retained rows packed into full micro-batches, takes about 40\% less training time, while rollout generation remains unchanged. Together, our measurements and training results connect outcome-conditioned supervision redundancy to more compute-efficient distillation.
}

\begin{document}
\maketitle

\section{Introduction}
\label{sec:1}

On-policy distillation trains a language model on trajectories sampled from its own policy, using a teacher's token-level distributions to supervise the prefixes the student actually visits \citep{ref1}. On-policy self-distillation (OPSD) obtains this supervision without a separately trained, larger teacher: the student sees a problem, while a teacher instantiated from the same base model additionally sees a reference solution \citep{ref2}. The reference supplies privileged information during training but is not required at inference. This creates a simple route to improving reasoning, but leaves an allocation question open: \emph{how much of this privileged supervision must actually be computed?}

Standard OPSD applies its distillation objective to every sampled trajectory, irrespective of whether the student reached the correct final answer. At the training pass rates in our experiments, roughly one-third of the rollouts are already correct. Each nevertheless incurs a student forward, a teacher forward, and student backward computation. Correctness does not make this supervision trivially unnecessary: a correct answer need not imply a correct derivation, and the teacher may still improve the student's intermediate distributions. The relevant question is empirical: can that supervision be omitted without sacrificing the improvement obtained by full OPSD?

Our gradient measurements motivate selective distillation: incorrect-response gradients are more coherent, while the two outcome populations largely share a mean direction.

Motivated by this observation, we propose \emph{correctness-filtered on-policy self-distillation} (C-OPSD). Our method generates one response per training problem, checks its final answer, and distills only a subset of incorrect responses. We physically remove unselected rows before both distillation forwards, reducing teacher/student forward and student backward work without changing rollout generation or the token-level distillation target.

Across Qwen3-1.7B, 4B, and 8B on AIME24, AIME25, and HMMT25, our method achieves \emph{on-par performance} with full OPSD while substantially reducing distillation computation: it executes approximately 68\% fewer rows in each teacher/student forward and student backward, and takes about 40\% less training time. We evaluate both the performance--computation trade-off and the role of outcome-based supervision allocation.

SCOPE and I-SDPO assign different training objectives according to rollout outcomes \citep{ref3,ref4}. We instead study whether substantial supervision can be \emph{removed} under a standalone OPSD objective while retaining performance, and investigate the gradient structure that makes this possible. This distinction motivates both our compute-aware implementation and our analysis of redundancy between outcome-conditioned gradient populations.

Our contributions are threefold:

\begin{itemize}
\item \textbf{Measurements of outcome-conditioned supervision redundancy.} We characterize gradient coherence and shared directions across correct and incorrect responses to explain opportunities for selective distillation.
\item \textbf{A measurement-motivated filtering method.} C-OPSD retains a subset of incorrect trajectories and removes unselected rows before both distillation forwards to avoid their computation.
\item \textbf{Performance and computation evaluation.} At three model scales (Qwen3-1.7B, 4B, and 8B), our method matches full OPSD (aggregate differences within \ensuremath{\pm}0.23 points) while executing 68\% fewer distillation rows and reducing training time by 40--41\% (14.2 vs.\ 8.4 minutes for 100 steps at 1.7B); allocation controls show that excluding correct trajectories, not retaining more incorrect ones, preserves accuracy.
\end{itemize}

\section{Background}
\label{sec:2}

\subsection{On-policy and privileged-context distillation}

Knowledge distillation transfers a teacher's output distribution to a student \citep{ref18}, and sequence-level distillation extends it to text generation \citep{ref19}. On-policy variants for large language models train on student-generated samples \citep{ref20,ref21}, following the imitation-learning insight that training on the learner's own state distribution reduces compounding errors \citep{ref22}.

In on-policy distillation, a student generates a response and a teacher provides token-level supervision along that response's prefixes. Unlike training exclusively on teacher-generated sequences, this places supervision on states visited by the student, including those arising from its own mistakes \citep{ref1}. The sampled tokens determine the prefixes at which distributions are compared; they are not themselves treated as ground-truth answers to imitate.

On-policy self-distillation (OPSD) obtains the teacher's informational advantage from privileged context rather than a separately trained, larger model \citep{ref2}. Distilling information available only in an extended context into a model that does not see it has also been studied as context distillation \citep{ref23}. Let (\emph{x}, \emph{r}) be a training problem and its reference solution, and let \emph{y} = (\emph{y}\textsubscript{1}, \ldots{}, \emph{y}\textsubscript{T}) be a student-generated response. At response position \emph{t}, the student and teacher distributions are

\begin{equation}
\begin{aligned}
p_{\theta,t} &= \pi_\theta(\cdot\mid x,y_{<t}),\\
q_{\bar\theta,t} &= \pi_{\bar\theta}(\cdot\mid x,r,y_{<t}).
\end{aligned}
\label{eq:1}
\end{equation}

Both models are evaluated on the same student-response prefix, but only the teacher sees the reference solution. The notation suppresses the respective prompt templates. In our fixed-teacher configuration, the student trains a LoRA adapter over a frozen base model, while the teacher uses that base model with the adapter disabled. Teacher weights therefore remain fixed as the student adapter changes. The teacher supplies distributions through forward evaluation; it does not generate a second response, and no gradient is propagated through its outputs.

\subsection{The OPSD training objective}

For a micro-batch \ensuremath{\mathcal{B}} of sampled responses, let \emph{m}\textsubscript{it} indicate a valid response-token position, excluding prompt and padding positions. Writing \emph{D}\textsubscript{\ensuremath{\tau}} for the implementation's clipped token-level divergence, the empirical OPSD objective is

\begin{equation}
\mathcal{L}_{\mathrm{OPSD}}(\theta;\mathcal{B})=
\frac{\displaystyle\sum_{i\in\mathcal{B}}\sum_t m_{it}
D_\tau\!\left(p_{\theta,it},\operatorname{sg}[q_{\bar\theta,it}]\right)}
{\displaystyle\max\!\left(1,\sum_{i\in\mathcal{B}}\sum_t m_{it}\right)}.
\label{eq:2}
\end{equation}

where sg denotes stop-gradient. Sampled responses are treated as fixed when differentiating this loss: the update differentiates the student distributions, not the discrete sampling procedure. We retain \emph{D}\textsubscript{\ensuremath{\tau}} as a named operator here because our intervention concerns trajectory selection, not the choice of distillation divergence.

Equation (2) averages over valid tokens within the micro-batch, rather than assigning equal total weight to every response. Responses of different lengths can consequently contribute different numbers of loss terms. The standard mask does not depend on final-answer correctness: correct and incorrect responses are supervised through the same token-level objective.

\subsection{Outcome correctness}

For each sampled response, let \emph{c}\textsubscript{i} \ensuremath{\in} \{0, 1\} be the outcome label returned by a symbolic answer checker using the training reference answer. This label becomes available after generation and does not require evaluating the privileged teacher. It is a property of the sampled response, not an intrinsic binary difficulty label for the problem: different attempts at the same problem may receive different outcomes. The checker evaluates the final answer rather than the intermediate reasoning steps.

Using this label to choose which trajectories receive distillation is distinct from using a binary reward in a policy-gradient objective. The supervision on a retained trajectory still comes from the teacher's conditional distributions. In particular, distilling an incorrect trajectory does not mean maximizing the likelihood of its incorrect final answer.

\subsection{Where the computation is spent}

An OPSD step contains three computational components: student rollout generation, student and teacher evaluation on the sampled trajectories, and backward computation through the student's loss. The teacher has no backward pass. Generation and subsequent distillation evaluation are separate costs.

A loss mask selects positions in the objective, whereas physical row selection removes trajectories from model evaluation. Selecting rows before the distillation forwards saves both forward evaluation and subsequent student backward work. We track supervision retention, executed forward/backward workload, and end-to-end runtime separately; C-OPSD reduces the distillation workload while retaining the baseline generation procedure.

\section{Related Work}
\label{sec:3}

\subsection{On-policy and privileged-context self-distillation}

On-policy distillation supervises prefixes generated by the student, addressing the distribution mismatch between teacher-generated training sequences and student behavior at inference \citep{ref1}. OPSD obtains teacher supervision from the same base model augmented with a reference solution \citep{ref2}. SDPO similarly converts additional context into dense self-distillation targets, using environment feedback or successful student attempts to inform the self-teacher \citep{ref16}. Its teacher template also reevaluates successful attempts using the attempt itself as the correct-solution context. These methods establish privileged-context supervision; C-OPSD keeps the OPSD teacher and token-level target fixed and studies which generated trajectories need that supervision computed.

\subsection{Correctness-dependent supervision}

SCOPE separates correct and incorrect rollouts, applies a student-side likelihood objective to correct responses, and uses teacher-guided distillation on incorrect responses, with perplexity-based weighting on both branches \citep{ref3}. I-SDPO routes groups of responses for the same question: all-incorrect groups with available ground truth use privileged self-distillation, whereas groups containing any success use group-relative reinforcement learning, including their incorrect responses \citep{ref4}. C-OPSD instead samples one rollout per question per sampling round, removes correct trajectories without assigning a replacement likelihood or reinforcement-learning loss, and retains only a subset of incorrect trajectories for the original OPSD objective.

Negative Sample Reinforcement provides a related precedent outside distillation: learning from negatively rewarded responses alone can be effective without explicit positive reinforcement of correct responses \citep{ref17}. Its update suppresses incorrect responses and redistributes probability according to the policy's existing preferences. C-OPSD instead matches privileged teacher distributions at the prefixes of retained incorrect responses. The shared motivation is selective use of outcome-labeled experience; the supervisory signal and update are different.

Reinforcement learning with verifiable rewards, such as GRPO \citep{ref24}, also uses outcome labels; DAPO filters out prompts whose sampled responses are all correct or all incorrect, since they provide no group-relative signal \citep{ref25}. Our method likewise discards outcome-labeled samples that add little signal, but does so within a distillation objective and removes the rows before computation.

\subsection{Selecting problems, trajectories, and tokens}

PACED estimates each problem's student pass rate from multiple samples and weights distillation toward intermediate competence, using a default kernel proportional to p(1 \ensuremath{-} p) \citep{ref14}. Most experiments estimate these weights once and keep them fixed; refresh schedules are also studied. This problem-level curriculum differs from filtering the outcome of a single newly sampled trajectory. A failed attempt on a usually solvable question remains eligible for C-OPSD, and the selection decision requires neither a multi-rollout pass-rate estimate nor teacher logits.

Two further distinctions are worth stating, because Equation (8) resembles PACED's kernel in form. That kernel is a function of each problem's estimated pass rate and concentrates weight at intermediate competence; Equation (8) is a function of the batch-level pass rate and describes what fraction of a batch survives, not which difficulty is preferred. The populations they select are in fact near opposite ends of the difficulty range: the mean offline pass rate of C-OPSD's retained set is 0.09 (Table~\ref{tab:7}), against the intermediate values PACED targets. PACED also reweights the loss, so every trajectory still passes through both forwards, whereas C-OPSD removes rows before them and converts the selection into computation saved.

At token level, TIP combines student entropy and teacher--student divergence to select informative positions, including confident predictions on which the teacher disagrees \citep{ref5}. STEPS identifies critical spans in student rollouts, applies self-distillation only to key or error spans, and combines it with GRPO \citep{ref15}. STEPS also transitions to pure GRPO after a short distillation phase, after which no teacher forwards are needed, providing a direct precedent for reducing teacher computation as well as selective supervision. C-OPSD uses a different intervention: whole trajectories are selected before both distillation forwards, removing discarded rows from student forward, teacher forward, and student backward computation without introducing span annotation or a replacement RL objective. Autoregressive generation is unchanged.

\citet{ref6} study the number and choice of distinct training problems. Their eight-hard-problem experiment reaches an average score of 53.6 versus 53.7 for the 17K-problem baseline. The selected problems are reused across training updates, with eight responses per prompt in the default setup; this is not an eight-rollout training budget. Their analysis links hard-problem benefits to longer reasoning trajectories. C-OPSD addresses a complementary efficiency axis: reducing distillation computation on already-generated online trajectories rather than reducing the number of distinct training problems.

Outside distillation, data-selection methods prioritize training points that are learnable and not yet learned \citep{ref26}, and selective language modeling trains only on tokens a reference model marks as useful \citep{ref27}.

\subsection{Gradient geometry and learning efficiency}

Gradient-coherence analyses characterize agreement among example-level updates \citep{ref7,ref8}, while gradient-noise scale connects stochastic-gradient variability to useful batch size \citep{ref9}. PACED applies a related signal--noise perspective to distillation: individual gradients on mastered problems need not vanish, but their dispersion across problems can yield low aggregate signal-to-noise ratio \citep{ref14}. Its empirical diagnostic bins problems by estimated pass rate and measures forward-KL gradients on fixed teacher-generated solutions.

Our measurements instead partition student-generated OPSD trajectories by observed correctness and jointly examine individual gradient magnitudes, within-population coherence, and overlap between the correct and incorrect population-mean directions. The additional cross-population comparison asks whether the more coherent incorrect trajectories also preserve the direction supplied by correct ones. We further test a fitted signal--noise model at larger group sizes without refitting and connect these measurements to a whole-trajectory compute-removal intervention. Routed OPD studies another level of the update, transporting probability mass between student-excess and teacher-deficit vocabulary alternatives at a given state \citep{ref10}. C-OPSD leaves that token-level distillation target unchanged and changes which trajectories are evaluated and differentiated.

\section{Outcome-Conditioned Gradient Geometry}
\label{sec:4}

We investigate which student-generated trajectories provide distinct distillation signal. The analysis separates three properties: individual gradient magnitude, coherence within each outcome population, and overlap between population-mean directions. These measurements motivate the allocation rule in Section 5; the training experiments in Section 6 evaluate its performance and computational consequences. Figure 1 summarizes the two main geometric findings examined in this section.

\begin{figure}[htbp]
\centering
\includegraphics[width=\textwidth,page=1,trim=31 171 10 62,clip]{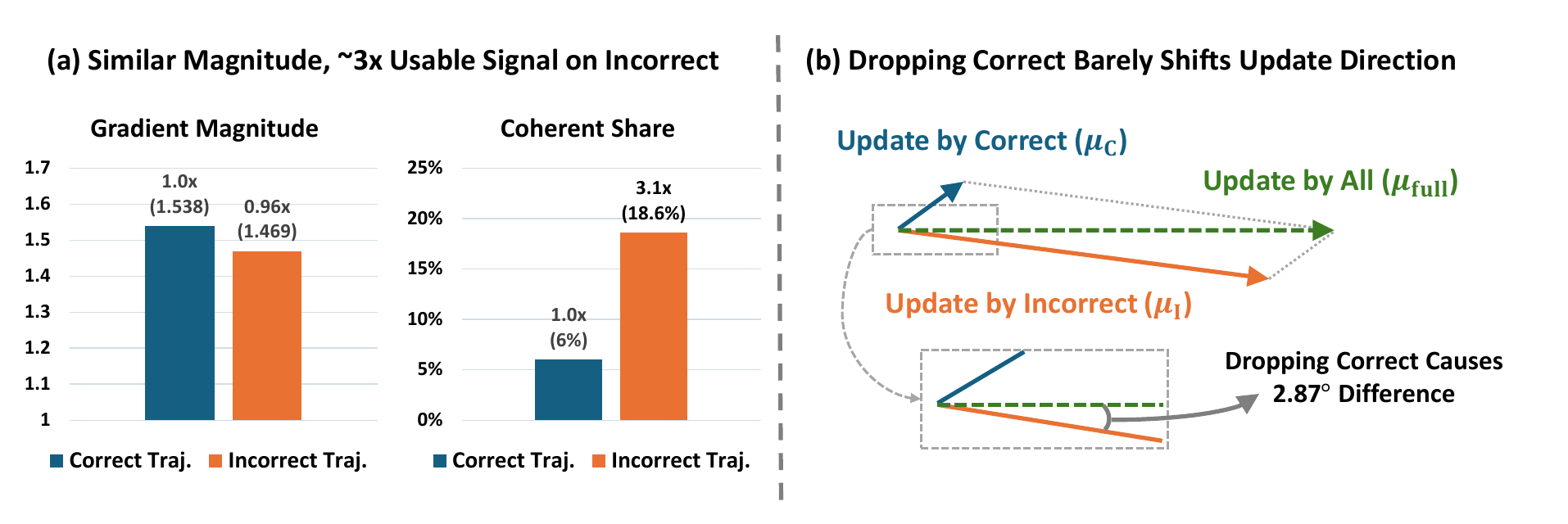}
\caption{\textbf{Outcome-conditioned gradient measurements motivate selective distillation.} Qwen3-1.7B at initialization, using 3,000 trajectories and base-weight gradients. \textbf{(a) Same cost, different return.} Individual gradient magnitudes are similar (1.538 vs.\ 1.469), but the fitted coherent share on incorrect trajectories is $3.1\times$ that on correct ones (18.6\% vs.\ 6.0\%), so incorrect trajectories preserve more usable signal per unit of distillation work. \textbf{(b) Dropping correct barely changes the update.} The response-weighted mixture $\mu_{\mathrm{full}} = p\mu_C + (1-p)\mu_I$, with $p = 0.291$, is only $2.87^{\circ}$ from the incorrect mean direction $\mu_I$. Vectors use measured mean magnitudes and the noise-corrected mean-direction cosine; the lower inset is an equal-axis $3\times$ enlargement of the tip region. Angles describe estimated population means. See \S\ref{sec:measurement} for procedure and derivation.}
\label{fig:1}
\end{figure}

\subsection{Measurement setup}
\label{sec:measurement}

We collect 3,000 trajectories from the Qwen3-1.7B training distribution at initialization: 873 correct and 2,127 incorrect responses, for a 29.1\% pass rate. Sampling uses non-thinking responses, a 1,024-token generation limit, temperature 1.1, top-p 0.95, and top-k 20.

Correct and incorrect responses also come from almost disjoint parts of the difficulty range (Table~\ref{tab:difficulty}): measured against an offline eight-rollout pass-rate probe, 62\% of correct responses fall on problems the base model solves at least seven times in eight, and 69\% of incorrect responses on problems it never solves.

\begin{table}[htbp]
\centering
\caption{Offline pass rate p\ensuremath{{}_0} of the problems behind the measured trajectories, computed on the 2,027 of 3,000 trajectories whose problems appear in the probe.}
\label{tab:difficulty}
\small
\setlength{\tabcolsep}{3pt}
\renewcommand{\arraystretch}{1.15}
\begin{tabularx}{\textwidth}{@{}*{5}{>{\raggedright\arraybackslash}X}@{}}
\toprule
\textbf{Population} & \textbf{Trajectories} & \textbf{Mean p\ensuremath{{}_0}} & \textbf{p\ensuremath{{}_0} = 0} & \textbf{p\ensuremath{{}_0} \ensuremath{\ge} 0.875} \\
\midrule
All & 2,027 & 0.34 & 45.8\% & 21.5\% \\
Correct & 680 & 0.80 & 0.0\% & 62.1\% \\
Incorrect & 1,347 & 0.11 & 68.9\% & 1.0\% \\
\bottomrule
\end{tabularx}
\end{table}

Within the middle band 0.125 \ensuremath{\le} p\ensuremath{{}_0} < 0.875 (258 correct and 406 incorrect trajectories), incorrect responses remain more coherent, with a coherent share of 19.3\% against 10.9\% for correct responses at group size four. We do not select on difficulty directly, because estimating p\ensuremath{{}_0} requires several rollouts per problem, whereas correctness requires a single answer check.

For each trajectory, we recompute the OPSD loss using a teacher context containing the reference solution and collect gradients with respect to the base-weight matrices targeted by LoRA. This measurement space avoids an initialization-specific degeneracy in adapter coordinates: when the B factor is initialized to zero, the gradient of the A factor also vanishes. Base-weight gradients let us compare the update directions that different trajectories induce on the same set of model weights.

We partition trajectories by final-answer correctness and construct four disjoint groups in each trial: two correct groups and two incorrect groups. After summing the gradients within each group, we compute cosine similarities between the two correct groups, between the two incorrect groups, and across outcomes.

The group size m is the \textbf{number of trajectories aggregated for analysis}, not the number of rollouts generated per problem. Within-population alignment measures how consistently trajectories recover a shared direction; cross-population alignment measures whether the two populations supply overlapping directions.

\subsection{Individual gradient magnitude does not distinguish the populations}

Individual gradient magnitudes are 1.538 for correct trajectories and 1.469 for incorrect trajectories. They are similar, and correct trajectories do not have smaller gradients. Their lower signal efficiency therefore cannot be explained simply by claiming that a correct answer makes the distillation gradient negligible.

The distinction therefore requires looking beyond per-trajectory magnitude. We next examine how much gradient energy survives aggregation along a common direction.

\subsection{A signal--noise model of gradient coherence}

Individual gradient magnitude does not fully characterize the learning signal. Large gradients can cancel when they point in different directions, whereas aligned gradients preserve a stronger common signal after aggregation.

For either outcome-conditioned population, write a per-trajectory gradient as

\[
g=\mu+\epsilon,\qquad\mathbb{E}[\epsilon]=0,\qquad\sigma^2=\mathbb{E}\lVert\epsilon\rVert^2.
\]

Here, \ensuremath{\mu} is the population-mean gradient and \ensuremath{\epsilon} is a trajectory's deviation from that mean. Noise refers to \textbf{across-trajectory gradient variability}, rather than incorrect answers or erroneous labels.

For the sum G\textsubscript{m} of m independently sampled gradients,

\begin{equation}
\mathbb{E}\lVert G_m\rVert^2=m^2\lVert\mu\rVert^2+m\sigma^2.
\label{eq:3}
\end{equation}

The shared-direction energy grows quadratically with m, while noise energy grows linearly. The common direction therefore becomes clearer as more trajectories are aggregated. When group norms concentrate sufficiently, the expected cosine between two independent groups from the same population is approximately

\begin{equation}
c_m\approx\frac{\lVert\mu\rVert^2}{\lVert\mu\rVert^2+\sigma^2/m}.
\label{eq:4}
\end{equation}

This relation connects measurable group alignment to the signal--noise structure of individual trajectories. We define the coherent share as

\[
q=\frac{\lVert\mu\rVert^2}{\lVert\mu\rVert^2+\sigma^2}\approx\frac{c_m}{m-(m-1)c_m}.
\]

Fitting the model to measurements at group size four gives the parameters in Table~\ref{tab:1}.

\begin{table}[htbp]
\centering
\caption{Signal--noise parameters fitted at group size four.}
\label{tab:1}
\small
\setlength{\tabcolsep}{3pt}
\renewcommand{\arraystretch}{1.15}
\begin{tabularx}{\textwidth}{@{}*{5}{>{\raggedright\arraybackslash}X}@{}}
\toprule
\textbf{Population} & \textbf{Within-population cosine} & \textbf{Mean-gradient magnitude \ensuremath{\Vert}\ensuremath{\mu}\ensuremath{\Vert}} & \textbf{Noise scale \ensuremath{\sigma}} & \textbf{Coherent share q} \\
\midrule
Correct & 0.2042 & 0.3776 & 1.4911 & 6.0\% \\
Incorrect & 0.4778 & 0.6339 & 1.3253 & 18.6\% \\
\bottomrule
\end{tabularx}
\end{table}

The incorrect population has approximately three times the coherent share of the correct population. The distinction is therefore not simply how large individual gradients are, but \textbf{how much of their energy survives aggregation along a common direction}.

The difference emerges after aggregation. Incorrect trajectories have a 1.68-fold larger mean-gradient magnitude and a lower noise scale. Defining per-trajectory signal-to-noise ratio as

\[
\mathrm{SNR}_1=\frac{\lVert\mu\rVert}{\sigma},
\]

the incorrect population has approximately 1.89 times the SNR of the correct population.

Under independent sampling from a fixed population, the SNR of m aggregated trajectories scales as \ensuremath{\surd}m. Correct trajectories would therefore require approximately 1.89\ensuremath{{}^2} \ensuremath{\approx} 3.6 times as many samples to attain the same mean-estimation SNR. This ratio describes \textbf{sample efficiency for estimating the mean gradient}.

This gives our first central finding: \textbf{correct trajectories produce substantial gradients, but those gradients are more dispersed; incorrect trajectories recover a stable common direction with fewer samples.} This difference motivates prioritizing incorrect trajectories when distillation computation is limited.

\subsection{Testing predictions at different group sizes}

If this decomposition captures the gradient structure, parameters estimated from small groups should also predict alignment in larger groups. We fix the parameters fitted at group size four and compare the resulting predictions with measurements at other group sizes, without refitting.

\begin{figure}[htbp]
\centering
\includegraphics[width=\textwidth,page=3,trim=82 146 82 87,clip]{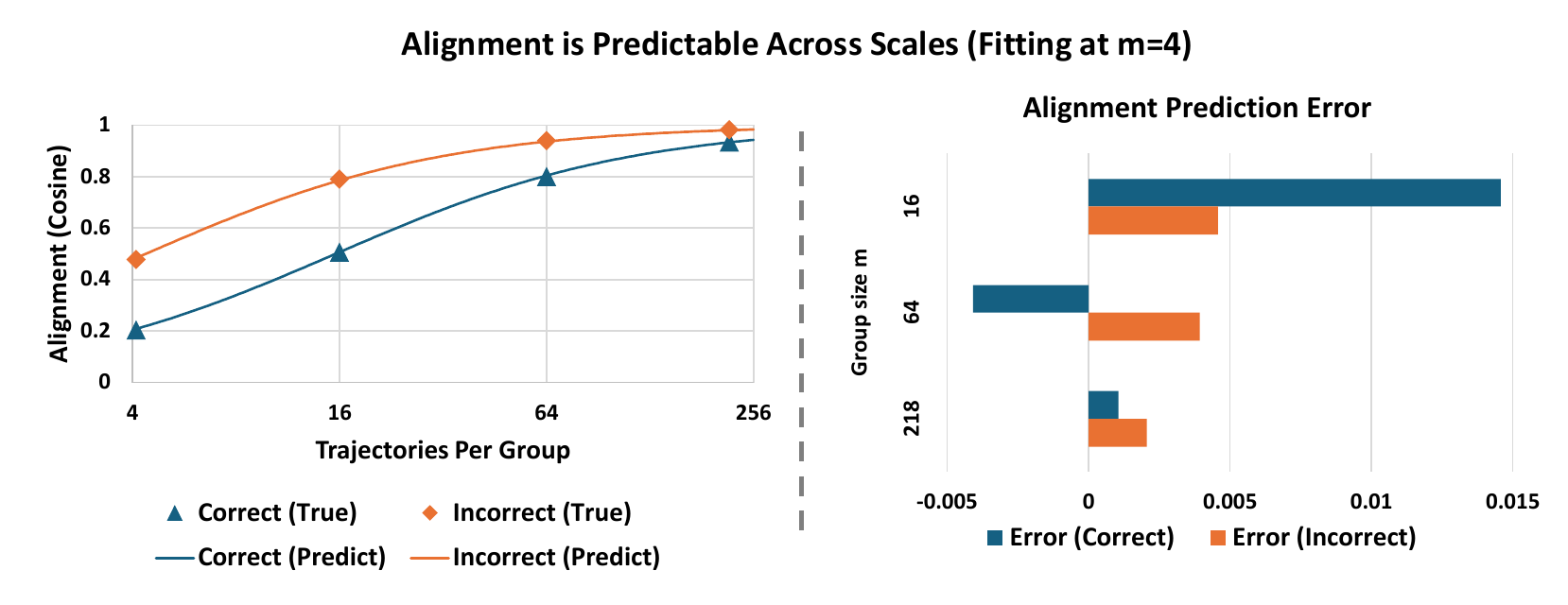}
\caption{\textbf{Alignment is predictable across scales from a fit at $m=4$.} \textbf{(Left)} Within-population gradient cosine vs.\ group size $m$ (trajectories per group, log-spaced ticks). Curves use $\|\mu\|,\sigma$ fitted at $m=4$ only ($N=64$ trials); markers at $m=16,64,218$ ($N=8,6,2$) are out-of-sample measurements (True vs.\ Predict), without refitting. \textbf{(Right)} Prediction error (measured $-$ predicted) on the same $m$; $m=4$ omitted (zero by construction). Max $|\mathrm{err}|=0.015$, RMSE $=0.007$. Group size counts trajectories aggregated for analysis, not rollouts per problem. All measurements are from Qwen3-1.7B at initialization.}
\label{fig:2}
\end{figure}

\begin{table}[htbp]
\centering
\caption{Predicted and measured within-population cosine at different group sizes.}
\label{tab:2}
\small
\setlength{\tabcolsep}{3pt}
\renewcommand{\arraystretch}{1.15}
\begin{tabularx}{\textwidth}{@{}*{5}{>{\raggedright\arraybackslash}X}@{}}
\toprule
\textbf{Group size} & \textbf{Correct: predicted} & \textbf{Correct: measured} & \textbf{Incorrect: predicted} & \textbf{Incorrect: measured} \\
\midrule
16 & 0.506 & 0.521 & 0.785 & 0.790 \\
64 & 0.804 & 0.800 & 0.936 & 0.940 \\
218 & 0.933 & 0.934 & 0.980 & 0.982 \\
\bottomrule
\end{tabularx}
\end{table}

Predictions closely track measurements across aggregation scales. The model captures both the increase in alignment with group size and the different rates at which the populations approach their mean directions: \textbf{incorrect trajectories yield a stable direction in smaller groups, whereas correct trajectories require more samples to average out variability.}

Large-group measurements provide a second cross-check. At group size 218, estimated mean-gradient magnitudes are approximately 0.375 and 0.633 for correct and incorrect trajectories, close to 0.378 and 0.634 from the small-group fit. These results support using the same signal--noise parameters to describe gradient behavior across aggregation scales.

\subsection{Stability during training}

We repeat the measurements on trajectories sampled from a C-OPSD checkpoint (1.7B) at steps 25, 50, and 100, using the same protocol at group size 16.

\begin{table}[htbp]
\centering
\caption{Coherent share and noise-corrected cross-population cosine during C-OPSD training, group size 16. The coherence ratio stays close to three and the two mean directions stay closely aligned at every checkpoint.}
\label{tab:training-stability}
\small
\setlength{\tabcolsep}{3pt}
\renewcommand{\arraystretch}{1.15}
\begin{tabularx}{\textwidth}{@{}*{5}{>{\raggedright\arraybackslash}X}@{}}
\toprule
\textbf{Checkpoint} & \textbf{Correct q} & \textbf{Incorrect q} & \textbf{Ratio} & \textbf{Corrected cross cosine} \\
\midrule
Initialization & 6.4\% & 19.1\% & 2.99 & 0.975 \\
Step 25 & 8.3\% & 25.9\% & 3.12 & 0.980 \\
Step 50 & 11.3\% & 33.2\% & 2.94 & 0.941 \\
Step 100 & 11.5\% & 33.9\% & 2.95 & 0.943 \\
\bottomrule
\end{tabularx}
\end{table}

Both populations become more coherent as training proceeds, but the ratio between them stays close to three, and the two mean directions remain closely aligned. The two findings that motivate C-OPSD hold throughout the regime in which it operates, not only at initialization.

\subsection{Shared mean directions and the effect of omitting correct responses}

Greater coherence alone does not explain why correct trajectories can be omitted. If they supply a distinct, important direction, dropping them could remove complementary supervision despite their higher noise. We therefore compare the two population-mean directions.

At group size 218, the measured cross-population cosine is 0.9267. Finite groups still contain sampling variability, so this value reflects both the difference between the mean directions and within-population noise. Using each population's internal alignment, a directional-noise approximation gives

\begin{equation}
\cos(\mu_{\mathrm{correct}},\mu_{\mathrm{incorrect}})
\approx\frac{0.9267}{\sqrt{0.9343\times0.9824}}\approx0.967.
\label{eq:5}
\end{equation}

The corresponding angle is approximately 14.7\ensuremath{{}^\circ}. Applying the same correction at group sizes 16, 64, and 218 gives 0.974, 0.967, and 0.967, respectively, indicating a stable estimate of mean-direction overlap across aggregation scales.

Decomposing the correct population's mean into components parallel and orthogonal to the incorrect mean gives approximate magnitudes

\[
\lVert\mu_{\mathrm{correct},\parallel}\rVert\approx0.3653,\qquad\lVert\mu_{\mathrm{correct},\perp}\rVert\approx0.0958.
\]

The correct mean therefore lies predominantly along a direction already supplied by incorrect trajectories, with a smaller orthogonal component.

This gives our second central finding: \textbf{correct trajectories largely reinforce an existing direction rather than supplying a separate one.} Together with the coherence analysis, this suggests that the reason to omit correct trajectories is not that their gradients are harmful, but that they provide noisier, substantially overlapping supervision.

To quantify the effect of removing the correct population from the response-weighted mean, let p = 873/3,000 = 0.291 and let \ensuremath{\theta} be the estimated angle between the two population means. The mixture is

\[
\mu_{\mathrm{full}}=p\mu_{\mathrm C}+(1-p)\mu_{\mathrm I}.
\]

Using \ensuremath{\Vert}\ensuremath{\mu}\textsubscript{C}\ensuremath{\Vert} = 0.3776, \ensuremath{\Vert}\ensuremath{\mu}\textsubscript{I}\ensuremath{\Vert} = 0.6339, and the unrounded noise-corrected cosine from Equation (5), the angle between this mixture and the incorrect mean is

\[
\varphi=\arctan\!\left(\frac{p\lVert\mu_{\mathrm C}\rVert\sin\theta}{(1-p)\lVert\mu_{\mathrm I}\rVert+p\lVert\mu_{\mathrm C}\rVert\cos\theta}\right)\approx2.87^\circ.
\]

The correct population is less frequent and has a smaller mean-gradient magnitude, limiting its contribution to the mixture's orthogonal component. The resulting mixture direction deviates from the incorrect mean by approximately 2.87\ensuremath{{}^\circ}, with a normalized orthogonal projection residual of approximately 5.0\%. Figure 1 illustrates this response-weighted geometry.

Together, greater incorrect-gradient coherence and overlapping mean directions motivate allocating less distillation work to correct responses. C-OPSD operationalizes this allocation by retaining a subset of incorrect responses, and the training experiments test whether the resulting reduction in supervision preserves performance.

\section{Method}
\label{sec:5}

Our gradient measurements motivate reducing supervision on correct responses. We implement this idea in C-OPSD through outcome verification after generation and physical row removal before distillation, as shown in Figure 3.

\begin{figure}[htbp]
\centering
\includegraphics[width=\textwidth,page=2,trim=15 27 39 22,clip]{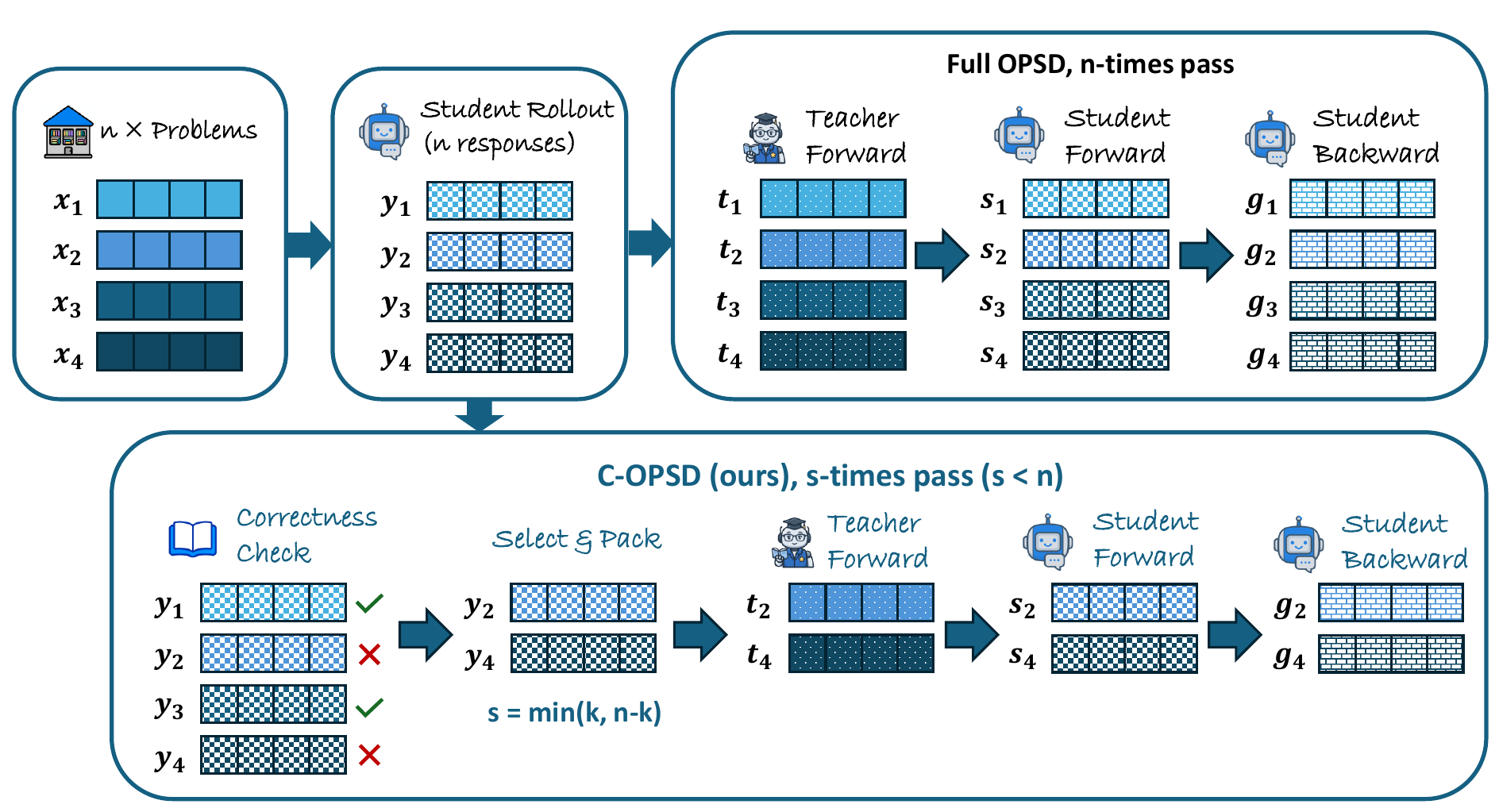}
\caption{\textbf{One rollout per problem with physically compacted distillation batches.} The student generates one response $y_i$ per problem $x_i$; generation is unchanged. Full OPSD (top right) runs teacher forward ($t_i$), student forward ($s_i$), and student backward ($g_i$) on all $n$ rows. C-OPSD (bottom) checks correctness against the training reference without teacher scoring, retains $s = \min(k, n-k)$ incorrect responses via Select \& Pack, and removes unselected rows before both distillation forwards, producing $s$-row tensors rather than $n$-row tensors with masked entries. Shown example: $n=4$ with $k=2$ correct ($y_1,y_3$), retaining $y_2,y_4$. The reference-conditioned teacher is frozen; only the student is updated. Empty selections execute one fully masked fallback row.}
\label{fig:3}
\end{figure}

\subsection{Outcome verification after generation}

Consider a micro-batch of n distinct training problems, each with one student-generated response. We retain the same student rollout procedure and privileged reference context as the OPSD baseline. Once generation has finished, an answer extractor and symbolic checker compare each response with its training reference answer, producing c\textsubscript{i} \ensuremath{\in} \{0, 1\}. The implementation uses the same math\_verify-based verification path described for evaluation, allowing equivalent forms such as 1/2 and 0.5 to agree. This decision uses training labels, not evaluation-benchmark answers.

Verification operates directly on the generated answer and training reference, enabling selection before either distillation forward without a second generation or teacher-scoring pass.

\subsection{Correctness-filtered retention}

Let C = \{i : c\textsubscript{i} = 1\} and W = \{i : c\textsubscript{i} = 0\}, with k = |C| and |W| = n \ensuremath{-} k. C-OPSD retains the first s = min(k, n \ensuremath{-} k) indices in W under the existing micro-batch order. Thus all retained trajectories are incorrect, but not every incorrect trajectory is retained. We denote the resulting index set by S.

\begin{equation}
s=\min(k,n-k),\qquad S=\operatorname{first}_s(W).
\label{eq:6}
\end{equation}

The dependence on k defines the retained budget used in the selection study. Because s is capped by the number of available incorrect responses, C-OPSD is not always exactly count-matched to a correct-only arm that retains all k correct responses. It is exactly count-matched, on the same sampled batch, to a random control selecting s responses regardless of correctness. Across separate on-policy training runs, sampled responses and hence realized counts can diverge.

For k = 0 or k = n, the rule selects no trajectory. For n = 1 it always selects none, so this particular budget rule requires larger micro-batches to provide training signal. The method's selection behavior depends on micro-batch size, not only on the total optimizer-step batch size.

\subsection{Retained objective and row removal}

For a retained set S, the filtered objective is the same token divergence as in Equation (2), normalized over the valid tokens of retained trajectories:

\begin{equation}
\mathcal{L}_{\mathrm{C\text{-}OPSD}}(\theta;\mathcal{B},S)=
\frac{\displaystyle\sum_{i\in S}\sum_t m_{it}
D_\tau\!\left(p_{\theta,it},\operatorname{sg}[q_{\bar\theta,it}]\right)}
{\displaystyle\max\!\left(1,\sum_{i\in S}\sum_t m_{it}\right)}.
\label{eq:7}
\end{equation}

Equation (7) averages the distillation loss over retained response tokens, with the selection treated as fixed during differentiation. For fixed sampled trajectories and per-token model outputs, it is equivalent to masking all unselected rows in the full-shaped batch: both formulations use the same retained-token numerator and denominator.

To realize computational savings, we index-select the retained rows from student input IDs and attention masks, teacher input IDs and attention masks, and labels before either distillation forward. Derived tensors are formed after selection. Prompt-length values that are scalars are not row-indexed. For an empty selection, the implementation keeps one row with all response labels masked so that the loss remains zero without passing a zero-row batch to unsupported kernels.

\subsection{Algorithm and controls}

\begin{quote}
\textbf{Algorithm 1. C-OPSD with physical row removal.}

\begin{enumerate}
\item Generate one response per problem with the current student policy; retain the generated tokens and the training references.
\item Extract and verify final answers, construct C and W, and compute s = min(|C|, |W|).
\item Select the first s incorrect rows under the existing batch order.
\item If s > 0, index-select the same rows from all row-aligned student, teacher, and label tensors. If s = 0, keep one fully masked fallback row.
\item Evaluate the student and reference-conditioned teacher on the selected prefixes; keep teacher outputs detached.
\item Compute the retained-token loss, backpropagate through the student, and follow the baseline's accumulation and optimizer-step procedure.
\end{enumerate}

\end{quote}

\begin{table}[htbp]
\centering
\caption{Supervision-allocation strategies.}
\label{tab:3}
\small
\setlength{\tabcolsep}{3pt}
\renewcommand{\arraystretch}{1.15}
\begin{tabularx}{\textwidth}{@{}*{4}{>{\raggedright\arraybackslash}X}@{}}
\toprule
\textbf{Arm} & \textbf{Retained trajectories} & \textbf{Count for the same batch} & \textbf{Role} \\
\midrule
Full OPSD & All trajectories & n & Performance reference \\
Correct-only & All correct trajectories & k & Opposite outcome allocation \\
C-OPSD & First s incorrect trajectories & s = min(k, n \ensuremath{-} k) & Proposed filtered objective \\
Random selection & Outcome-blind subset of all trajectories & s & Count-matched selection control \\
All-incorrect & All incorrect trajectories & n \ensuremath{-} k & Higher-retention comparison \\
\bottomrule
\end{tabularx}
\end{table}

\subsection{Retention as a function of micro-batch size}

For an illustrative model with independent, identically distributed correctness labels of success probability p, K follows Binomial(n, p). The expected loss-retention fraction is r\textsubscript{n}(p) = E[min(K, n \ensuremath{-} K)]/n. For n = 4, direct enumeration gives:

\begin{equation}
r_4(p)=p(1-p)(1+p-p^2).
\label{eq:8}
\end{equation}

The maximum is 0.3125 at p = 0.5. At p = 0.33, the expected loss-retention fraction is approximately 0.270. Empty selections occur with probability p\textsuperscript{n} + (1 \ensuremath{-} p)\textsuperscript{n}. Under the one-row fallback, the expected executed-row fraction is r\textsubscript{n}(p) + [p\textsuperscript{n} + (1 \ensuremath{-} p)\textsuperscript{n}]/n. For n = 4 and p = 0.33 this is approximately 0.323, or 1.29 physical rows per micro-batch. For n = 2, the same implementation always executes one row: its physical retention is 50\%, even though its expected loss retention is p(1 \ensuremath{-} p).

The rule has no free parameter, but it does have a domain. Equation (8) makes r\textsubscript{4} a function of the pass rate: it peaks at 31.3\% near p = 0.5 and is flat over p \ensuremath{\in} [0.3, 0.5], which covers the 29--35\% pass rates observed in our runs. Outside that band it falls away on both sides, to approximately 22\% at p = 0.25, 18.6\% at p = 0.2, and 9.8\% at p = 0.1. Because the pass rate shifts as the student trains, from 33\% to 29\% over 100 steps at 1.7B, the budget also adapts without a schedule. A fixed-rate subsampling rule would be insensitive to this; min(k, n \ensuremath{-} k) trades that robustness for the absence of a tuned retention hyperparameter, and at our pass rates the two retain near-identical sets on the same sampled batches. We therefore state the rule as intended for pass rates in roughly [0.25, 0.75], and recommend a fixed retention fraction outside that range.

\section{Experimental Results}
\label{sec:6}

\subsection{Models, training, and evaluation}

We evaluate Qwen3 models \citep{ref13} at 1.7B, 4B, and 8B parameters using the OPSD training split named siyanzhao/Openthoughts\_math\_30k\_opsd. Our offline difficulty probe covers 29,427 problems. We use rank-64 LoRA \citep{ref11} with scaling \ensuremath{\alpha} = 128, targeting attention q, k, v, and o projections and MLP gate, up, and down projections. The reference-conditioned teacher uses the frozen base with the adapter disabled.

\begin{table}[htbp]
\centering
\caption{Training and evaluation settings.}
\label{tab:4}
\small
\setlength{\tabcolsep}{3pt}
\renewcommand{\arraystretch}{1.15}
\begin{tabularx}{\textwidth}{@{}*{2}{>{\raggedright\arraybackslash}X}@{}}
\toprule
\textbf{Setting} & \textbf{Value} \\
\midrule
Optimizer-step rollout count & 32 distinct problems; one response per problem \\
Student rollout mode & Non-thinking; vLLM \citep{ref12} colocated generation \\
Maximum rollout length & 1,024 tokens \\
Rollout sampling & Temperature 1.1; top-p 0.95; top-k 20 \\
Learning rate & \ensuremath{5\times 10^{-6}}; constant schedule \\
Token-divergence clip parameter & 0.05 for 1.7B/4B; 0.06 for 8B \\
Main checkpoint selection set & Steps 25, 50, 75, and 100 \\
Training runs per arm & Five (main comparison and ablation); three (SFT, GRPO) \\
Evaluation & Thinking mode; maximum 38,912 new tokens; temperature 1.0; top-p 0.95 \\
\bottomrule
\end{tabularx}
\end{table}

\begin{table}[htbp]
\centering
\caption{Batch configurations across experiments.}
\label{tab:5}
\small
\setlength{\tabcolsep}{3pt}
\renewcommand{\arraystretch}{1.15}
\begin{tabularx}{\textwidth}{@{}*{5}{>{\raggedright\arraybackslash}X}@{}}
\toprule
\textbf{Result block} & \textbf{Per-device batch} & \textbf{Gradient accumulation} & \textbf{Ranks} & \textbf{Total rollouts/step} \\
\midrule
All experiments & 4 & 8 & 1 & 32 \\
\bottomrule
\end{tabularx}
\end{table}

For each benchmark problem, Avg@M averages correctness over M sampled responses and then averages over problems, measuring sampled-answer accuracy. All reported results use Avg@48. Each of the three benchmark grids contains 30 problems, yielding 1,440 sampled responses per benchmark. The main comparison and the outcome-allocation ablation use five independent training runs per arm; SFT and GRPO use three. Rollout sampling is not seeded, so repeated runs of one configuration diverge on their own; training jobs also vary the trainer seed, which changes the data order. Reported scores are means over runs.

The main table uses the best checkpoint for each benchmark among steps 25, 50, 75, and 100; its aggregate combines those benchmark-specific selections. We also report per-benchmark accuracy at each of these checkpoints using the same runs. The main selection uses only the first 100 steps, including for 1.7B runs that continued to step 200.

\subsection{Main comparison: on-par performance with less distillation}

Table~\ref{tab:6} reports all five arms at the three model scales. It contains two kinds of comparison. Both OPSD arms lead the untrained base model and the SFT and GRPO baselines on every benchmark at every scale, by margins that are large relative to the run-to-run standard deviations in the SD column. Between the two OPSD arms the aggregate differences (C-OPSD minus OPSD) are \ensuremath{-}0.19, +0.23, and +0.09 percentage points at 1.7B, 4B, and 8B: inconsistent in sign and of the same size as that variation, with 95\% confidence intervals of [\ensuremath{-}0.67, 0.28], [\ensuremath{-}0.35, 0.82], and [\ensuremath{-}0.24, 0.42]. At a fixed final checkpoint (step 100), C-OPSD minus OPSD is +0.41, +1.02, and +0.10. Our method is therefore on par with full OPSD, and retains the whole of its advantage over the baselines, while computing the distillation loss on roughly a quarter of the generated trajectories.

\begin{table}[htbp]
\centering
\caption{Best checkpoint per benchmark, Avg@48. Scores are percentages, averaged over training runs: five for the OPSD arms, three for SFT and GRPO, one for the untrained base. SD is the standard deviation of the per-run aggregate. A rule separates the baselines from the two OPSD arms within each model size. Retention is the fraction of generated responses distilled. Executed-row measurements are presented separately in Section 7.1. Our method matches full OPSD at every scale (aggregate differences \ensuremath{-}0.19, +0.23, and +0.09 points) while distilling only 26--29\% of generated responses.}
\label{tab:6}
\small
\setlength{\tabcolsep}{4pt}
\renewcommand{\arraystretch}{1.15}
\begin{tabular*}{\textwidth}{@{\extracolsep{\fill}}llrrrrrr@{}}
\toprule
\textbf{Model} & \textbf{Method} & \textbf{AIME24} & \textbf{AIME25} & \textbf{HMMT25} & \textbf{Mean} & \textbf{SD} & \textbf{Retention} \\
\midrule
1.7B & Base & 49.10 & 33.89 & 23.26 & 35.42 & --- & --- \\
 & SFT & 50.42 & 38.59 & 24.44 & 37.82 & 0.19 & 100\% \\
 & GRPO & 49.26 & 36.16 & 23.66 & 36.36 & 0.42 & 100\% \\
\cmidrule(lr){2-8}
 & OPSD & \textbf{56.19} & \textbf{41.85} & \textbf{28.74} & \textbf{42.26} & 0.29 & 100\% \\
 & C-OPSD & 56.18 & 41.50 & 28.51 & 42.06 & 0.34 & \textbf{26\%} \\
\midrule
4B & Base & 74.38 & 65.35 & 43.75 & 61.16 & --- & --- \\
 & SFT & 72.78 & 66.76 & 43.12 & 60.89 & 0.32 & 100\% \\
 & GRPO & 74.35 & 66.48 & 43.84 & 61.56 & 0.30 & 100\% \\
\cmidrule(lr){2-8}
 & OPSD & 74.96 & 67.90 & \textbf{45.03} & 62.63 & 0.47 & 100\% \\
 & C-OPSD & \textbf{75.54} & \textbf{68.47} & 44.57 & \textbf{62.86} & 0.18 & \textbf{29\%} \\
\midrule
8B & Base & 76.81 & 69.38 & 45.28 & 63.82 & --- & --- \\
 & SFT & 75.05 & 68.43 & 44.95 & 62.81 & 0.30 & 100\% \\
 & GRPO & 76.64 & 69.91 & 45.67 & 64.07 & 0.09 & 100\% \\
\cmidrule(lr){2-8}
 & OPSD & \textbf{77.50} & \textbf{70.72} & 46.65 & 64.96 & 0.20 & 100\% \\
 & C-OPSD & 77.43 & 70.67 & \textbf{47.06} & \textbf{65.05} & 0.24 & \textbf{28\%} \\
\bottomrule
\end{tabular*}
\end{table}

Appendix A (Table A1) provides the per-dataset, per-checkpoint breakdown for the two OPSD arms.

\subsection{Outcome-allocation ablation}

We next compare which trajectories receive a restricted budget. The 1.7B ablation uses the same protocol as the main comparison -- best checkpoint per benchmark, Avg@48 -- with five training runs per arm. Measured retention is 26\% for random selection and C-OPSD, so the random control is closely count-matched; correct-only retains 33\% and all-incorrect 68\%. Table~\ref{tab:7} reports the result.

\begin{table}[htbp]
\centering
\caption{Outcome allocation: best checkpoint per benchmark, Avg@48, mean over five training runs. Offline p\ensuremath{{}_0} estimates the retained problems' pass rates from a separate eight-rollout probe. Full OPSD, at 100\% retention, scores 42.26 under the same protocol. Excluding correct trajectories is what matters: correct-only falls 1.71 points below our method, while all-incorrect, keeping 68\% of rows, scores no higher.}
\label{tab:7}
\small
\setlength{\tabcolsep}{3pt}
\renewcommand{\arraystretch}{1.15}
\begin{tabularx}{\textwidth}{@{}*{5}{>{\raggedright\arraybackslash}X}@{}}
\toprule
\textbf{Arm} & \textbf{Retention} & \textbf{Incorrect among retained} & \textbf{Mean p\ensuremath{{}_0}} & \textbf{Mean accuracy (SD)} \\
\midrule
Correct-only & 33\% & 0\% & 0.83 & 40.35 (0.20) \\
Random selection & 26\% & 66\% & 0.42 & 41.58 (0.31) \\
C-OPSD & 26\% & 100\% & 0.09 & \textbf{42.06} (0.34) \\
All-incorrect & 68\% & 100\% & 0.09 & 42.17 (0.38) \\
\bottomrule
\end{tabularx}
\end{table}

The pattern is consistent across the four allocations. Correct-only is by far the weakest, 1.71 points below C-OPSD: spending the budget on trajectories the student already solved is the one choice that clearly costs accuracy. Random selection sits between the two, 0.48 below C-OPSD; it already directs 66\% of its budget to incorrect responses, since those are the majority of generated trajectories, so restricting the remainder to incorrect responses moves the score only a little further. All-incorrect, which keeps 68\% of rollouts rather than 26\%, scores level with C-OPSD and with full OPSD, so the gain comes from excluding correct trajectories rather than from retaining more of the incorrect ones.

The retained populations also differ in problem difficulty: their mean offline pass rates are 0.83, 0.42, and 0.09 for correct-only, random, and C-OPSD, respectively. The ablation compares these outcome-conditioned populations, with response correctness and problem difficulty varying together.

\subsection{Gradient geometry and supervision allocation}

Across the three allocation strategies, the estimated shared-direction SNR follows the same ordering as measured accuracy: C-OPSD > Random selection > Correct-only. Table~\ref{tab:8} compares the SNR estimates, computed from the fitted gradient parameters and retained trajectory counts, with the best-checkpoint, Avg@48 ablation results.

\begin{table}[htbp]
\centering
\caption{Estimated SNR and measured ablation results for three supervision-allocation strategies. Estimated SNR orders the three arms exactly as accuracy does.}
\label{tab:8}
\small
\setlength{\tabcolsep}{3pt}
\renewcommand{\arraystretch}{1.15}
\begin{tabularx}{\textwidth}{@{}*{6}{>{\raggedright\arraybackslash}X}@{}}
\toprule
\textbf{Strategy} & \textbf{Expected retained trajectories/step} & \textbf{Shared-direction signal} & \textbf{Noise} & \textbf{Estimated SNR} & \textbf{Mean accuracy} \\
\midrule
Correct-only & 10.56 & 3.86 & 4.85 & 0.80 & 40.35 \\
Random selection & 8.32 & 4.51 & 3.99 & 1.13 & 41.58 \\
C-OPSD & 8.32 & 5.27 & 3.82 & 1.38 & 42.06 \\
\bottomrule
\end{tabularx}
\end{table}

The estimated SNR ordering is consistent with the observed accuracy ordering: C-OPSD exceeds random selection, which exceeds correct-only selection. C-OPSD and random selection retain nearly the same number of trajectories. Correct-only selection retains more trajectories yet has lower estimated SNR and lower measured accuracy.

The comparison requires similar retained counts, because aggregated SNR grows with the square root of that count. Full OPSD retains all 32 and scores 2.29 against C-OPSD's 1.38, yet the two are on par in Table~\ref{tab:6}. Table~\ref{tab:8} therefore speaks to \emph{which} trajectories are kept, not how many.

Together with the main comparison, these results connect the outcome-conditioned geometry to a practical allocation of distillation work. The geometry motivates the method; the training experiments show that C-OPSD achieves on-par performance with substantially less forward and backward computation.

\section{Computational Efficiency}
\label{sec:7}

\subsection{Measured forward/backward reduction}

C-OPSD converts trajectory filtering into actual computation savings by removing unselected rows before both distillation forwards. In the measured micro-batch-size-four implementation, each student forward, teacher forward, and student backward processes an average of 1.29 rows instead of 4, a reduction of approximately 68\%. The count includes the fully masked fallback row executed for an empty selection. The retained-token objective is defined in Section 5.3.

\begin{table}[htbp]
\centering
\caption{Executed distillation rows at micro-batch size four. Our method executes about 68\% fewer rows in every distillation operation.}
\label{tab:9}
\small
\setlength{\tabcolsep}{3pt}
\renewcommand{\arraystretch}{1.15}
\begin{tabularx}{\textwidth}{@{}*{4}{>{\raggedright\arraybackslash}X}@{}}
\toprule
\textbf{Distillation operation} & \textbf{Full OPSD rows} & \textbf{C-OPSD mean rows} & \textbf{Executed-row reduction} \\
\midrule
Student forward & 4 & 1.29 & \ensuremath{\approx}68\% \\
Teacher forward & 4 & 1.29 & \ensuremath{\approx}68\% \\
Student backward & 4 & 1.29 & \ensuremath{\approx}68\% \\
\bottomrule
\end{tabularx}
\end{table}

All four student trajectories are generated before selection. The reduction therefore applies to distillation evaluation and backward work; Section 7.2 incorporates the unchanged generation cost into the total-step FLOP model.

\subsection{Estimated total-step FLOP savings}

For an illustrative model of dense parameter-matrix work, let N be the non-embedding parameter count, P the student prompt length, C the completion length, L\textsubscript{ref} the extra teacher-reference length, and R the number of physical rows used by distillation in an optimizer step. We estimate these components as follows:

\begin{equation}
\begin{aligned}
F_{\mathrm{generation}} &\approx 32(P+C)\times 2N,\\
F_{\mathrm{student}} &\approx R(P+C)\times 6N,\\
F_{\mathrm{teacher}} &\approx R(P+L_{\mathrm{ref}}+C)\times 2N.
\end{aligned}
\label{eq:9}
\end{equation}

The student term assumes a frozen-base LoRA implementation with activation-gradient work and one checkpoint-recomputation forward, rather than full base-weight gradient computation. For this estimate, we use P \ensuremath{\approx} 150, C = 978, L\textsubscript{ref} = 616, and change R from 32 to 10.32 for the n = 4 implementation. With these shared assumptions, the fractional saving is independent of N and is approximately 55.5\% of the modeled total step FLOPs.

\begin{table}[htbp]
\centering
\caption{Modeled FLOPs per optimizer step (TFLOPs), using the shared n = 4 configuration and lengths in Equation (9). Our method reduces modeled step FLOPs by 55.5\% at every scale, including the unchanged generation cost; the saving is identical across scales because every term in Equation (9) scales with N. Measured wall-clock savings (40--41\%) are reported in Table~\ref{tab:wallclock}.}
\label{tab:10}
\small
\setlength{\tabcolsep}{3pt}
\renewcommand{\arraystretch}{1.15}
\begin{tabular*}{\textwidth}{@{\extracolsep{\fill}}lrrrrr@{}}
\toprule
\textbf{\shortstack[l]{Model\\scale}} & \textbf{\shortstack[l]{Generation}} & \textbf{\shortstack[l]{Student\\f+b}} & \textbf{\shortstack[l]{Teacher\\f}} & \textbf{\shortstack[l]{Full\\total}} & \textbf{\shortstack[l]{Filtered\\total}} \\
\midrule
1.7B & 101.7 & 305.2 & 157.3 & 564.1 & 250.9 \\
4B & 262.3 & 786.8 & 405.5 & 1454.6 & 646.8 \\
8B & 501.5 & 1504.3 & 775.3 & 2781.1 & 1236.6 \\
\bottomrule
\end{tabular*}
\end{table}

The total-step FLOP estimate includes unchanged rollout generation and reduced distillation forward/backward work.

\subsection{Measured wall-clock time}

We measure training time on a single GB300 GPU with colocated vLLM and 32 rollouts per optimizer step (per-device batch 4, gradient accumulation 8). For both methods, all rollouts of a step are generated in one vLLM call, and to turn the row reduction into time, C-OPSD packs the retained rows of a step into full micro-batches of four. Both changes are mathematically equivalent to the original update, identical up to floating-point rounding, which we verified numerically (Appendix~\ref{sec:appendix-b}). Table~\ref{tab:wallclock} reports the wall-clock time of a full 100-step training loop, including checkpoint saves every 25 steps, and the median time per step over steps 2--12.

\begin{table}[htbp]
\centering
\caption{Measured wall-clock time on a single GB300 GPU. Our method reduces training time by 40--41\% relative to full OPSD at every scale.}
\label{tab:wallclock}
\small
\setlength{\tabcolsep}{3pt}
\renewcommand{\arraystretch}{1.15}
\begin{tabular*}{\textwidth}{@{\extracolsep{\fill}}lrrrrr@{}}
\toprule
& \multicolumn{3}{c}{\textbf{100-step training (min)}} & \multicolumn{2}{c}{\textbf{Per step (s)}} \\
\cmidrule(lr){2-4}\cmidrule(lr){5-6}
\textbf{Model} & \textbf{OPSD} & \textbf{C-OPSD} & \textbf{Saving} & \textbf{OPSD} & \textbf{C-OPSD} \\
\midrule
1.7B & 14.2 & 8.4 & 41\% & 7.86 & 4.79 \\
4B & 19.8 & 11.8 & 40\% & 12.36 & 6.95 \\
8B & 22.6 & 13.7 & 40\% & 12.98 & 8.05 \\
\bottomrule
\end{tabular*}
\end{table}

\section{Conclusion}
\label{sec:8}

Outcome-conditioned gradient measurements reveal a form of supervision redundancy in OPSD. Correct and incorrect responses have similar individual gradient magnitudes, but incorrect responses have greater coherence, while the two population-mean directions largely overlap. A signal--noise model fitted at a small group size tracks alignment at larger sizes, and the measured population proportions and mean magnitudes explain why omitting the correct population changes the estimated response-weighted mean direction only slightly.

Motivated by these findings, we propose C-OPSD, which generates one response per problem, retains a subset of incorrect responses, and removes unselected rows before both distillation forwards. Across Qwen3-1.7B, 4B, and 8B on AIME24, AIME25, and HMMT25, our method matches full OPSD. In a measured micro-batch-size-four implementation, executed rows decrease from 4 to 1.29 per teacher/student forward and student backward, a reduction of approximately 68\%, and packing the retained rows into full micro-batches turns this into about 40\% less training time. Our method preserves the token-level distillation target while concentrating computation on a smaller set of trajectories. Together, the measurements and training results demonstrate the practical value of allocating distillation work according to outcome-conditioned gradient structure.

\clearpage

\appendix

\section{Per-Dataset Checkpoint Results}

\label{sec:appendix-a}

\setcounter{table}{0}

\renewcommand{\thetable}{A\arabic{table}}

Table A1 reports accuracy on each benchmark at steps 25, 50, 75, and 100 for the same runs, averaged over the five training runs per arm. This breakdown shows checkpoint-level performance on AIME24, AIME25, and HMMT25 alongside the best-checkpoint comparison in Table~\ref{tab:6}.

\begin{table}[htbp]
\centering
\caption{Per-dataset accuracy (Avg@48) at each evaluation checkpoint, averaged over the five training runs per arm used in Table~\ref{tab:6}. Every benchmark peaks at step 50 or later, most often at step 75 or 100, for both methods.}
\label{tab:A1}
\small
\setlength{\tabcolsep}{3pt}
\renewcommand{\arraystretch}{1.15}
\begin{tabular*}{\textwidth}{@{\extracolsep{\fill}}lllrrrr@{}}
\toprule
\textbf{\shortstack[l]{Model}} & \textbf{\shortstack[l]{Dataset}} & \textbf{\shortstack[l]{Method}} & \textbf{\shortstack[l]{Step\\25}} & \textbf{\shortstack[l]{Step\\50}} & \textbf{\shortstack[l]{Step\\75}} & \textbf{\shortstack[l]{Step\\100}} \\
\midrule
1.7B & AIME24 & OPSD & 53.33 & 54.76 & 56.00 & 55.10 \\
1.7B & AIME24 & C-OPSD & 52.64 & 54.03 & 56.04 & 55.14 \\
\addlinespace[2pt]
1.7B & AIME25 & OPSD & 39.60 & 41.32 & 41.01 & 39.75 \\
1.7B & AIME25 & C-OPSD & 39.03 & 40.40 & 41.38 & 40.38 \\
\addlinespace[2pt]
1.7B & HMMT25 & OPSD & 24.57 & 27.08 & 28.74 & 27.89 \\
1.7B & HMMT25 & C-OPSD & 24.43 & 25.42 & 27.49 & 28.46 \\
\midrule
4B & AIME24 & OPSD & 74.03 & 74.93 & 74.53 & 73.72 \\
4B & AIME24 & C-OPSD & 74.56 & 74.85 & 75.49 & 75.03 \\
\addlinespace[2pt]
4B & AIME25 & OPSD & 67.11 & 67.42 & 67.64 & 66.14 \\
4B & AIME25 & C-OPSD & 66.57 & 67.51 & 68.04 & 67.54 \\
\addlinespace[2pt]
4B & HMMT25 & OPSD & 42.71 & 43.99 & 44.90 & 43.78 \\
4B & HMMT25 & C-OPSD & 42.60 & 42.83 & 44.17 & 44.14 \\
\midrule
8B & AIME24 & OPSD & 76.15 & 77.22 & 77.18 & 76.92 \\
8B & AIME24 & C-OPSD & 76.96 & 76.56 & 76.81 & 77.38 \\
\addlinespace[2pt]
8B & AIME25 & OPSD & 69.11 & 70.08 & 69.96 & 70.46 \\
8B & AIME25 & C-OPSD & 69.72 & 69.42 & 69.67 & 70.15 \\
\addlinespace[2pt]
8B & HMMT25 & OPSD & 45.43 & 45.89 & 46.36 & 46.47 \\
8B & HMMT25 & C-OPSD & 45.68 & 45.71 & 46.42 & 46.61 \\
\bottomrule
\end{tabular*}
\end{table}

\section{Implementation Details}
\label{sec:appendix-b}

\paragraph{Divergence.} With teacher distribution $q$ and student distribution $p$ at a response position, both computed from logits divided by the sampling temperature 1.1,
\[
D_\tau(p,q)=\sum_{v}\min\!\big(q_v\log(q_v/p_v),\,\tau\big),
\]
the forward KL with each vocabulary entry's contribution clipped at $\tau$ (0.05 for 1.7B and 4B, 0.06 for 8B), as in the OPSD implementation. The loss averages $D_\tau$ over valid response tokens.

\paragraph{Row packing.} All but the last micro-batch of an accumulation window apply the selection rule and pass their retained rows on. The last micro-batch packs all retained rows into micro-batches of four, left-padded to a common prompt width. Each token is weighted by $1/N_j$, where $N_j$ is the number of retained tokens in its original micro-batch $j$, so the accumulated gradient is the same as without packing. A micro-batch with no retained rows contributes nothing.

\paragraph{Baselines.} SFT and GRPO follow the setup of the OPSD paper~\citep{ref2}, with the same 100-step budget and checkpoint selection as the OPSD arms.

\paragraph{Gradient measurements.} The parameters in Table~\ref{tab:1} average 64 trials at group size 4; the out-of-sample measurements average 8, 6, and 2 trials at group sizes 16, 64, and 218; each training-stability measurement averages 8 trials at group size 16. Every trial draws four disjoint groups.

\end{document}